\documentclass[sigconf]{acmart}

\usepackage{graphicx}
\graphicspath{{figures/}}
\usepackage{algorithm}
\usepackage{algorithmic}
\usepackage{booktabs}
\usepackage{siunitx}
\usepackage{float}
\usepackage{xcolor}
\copyrightyear{2026}
\acmYear{2026}
\setcopyright{cc}
\setcctype{by-nc-nd}
\acmConference[KDD '26]{Proceedings of the 32nd ACM SIGKDD Conference on Knowledge Discovery and Data Mining V.1}{August 09--13, 2026}{Jeju Island, Republic of Korea}
\acmBooktitle{Proceedings of the 32nd ACM SIGKDD Conference on Knowledge Discovery and Data Mining V.1 (KDD '26), August 09--13, 2026, Jeju Island, Republic of Korea}
\acmPrice{}
\acmDOI{10.1145/3770854.3783929}
\acmISBN{979-8-4007-2258-5/2026/08}

\begin{document}

\title{KAN-FIF: Spline-Parameterized Lightweight Physics-based Tropical Cyclone Estimation on Meteorological Satellite}

\author{Jiakang Shen}
\orcid{0009-0008-7067-463X}
\affiliation{
  \institution{Shandong University}
  \department{School of Control Science and Engineering}
  \city{Jinan}
  \state{Shandong}
  \country{China}}
\email{202300171054@mail.sdu.edu.cn}

\author{Qinghui Chen}
\orcid{0000-0001-5504-2951}
\affiliation{
  \institution{Shandong University}
  \department{School of Control Science and Engineering}
  \city{Jinan}
  \state{Shandong}
  \country{China}}
\email{202420785@mail.sdu.edu.cn}

\author{Runtong Wang}
\orcid{0009-0002-1663-3649}
\affiliation{
  \institution{Shandong University}
  \department{School of Control Science and Engineering}
  \city{Jinan}
  \state{Shandong}
  \country{China}}
\email{202300171170@mail.sdu.edu.cn}
  
\author{Chenrui Xu}
\orcid{0009-0004-9563-9065}
\affiliation{
  \institution{Shandong University}
  \department{School of Control Science and Engineering}
  \city{Jinan}
  \state{Shandong}
  \country{China}}
\email{202300171055@mail.sdu.edu.cn}

\author{Jinglin Zhang}
\authornote{Corresponding author}
\orcid{0000-0003-1618-8493}
\affiliation{
  \institution{Shandong University}
  \department{School of Control Science and Engineering}
  \city{Jinan}
  \state{Shandong}
  \country{China}}
\email{jinglin.zhang@sdu.edu.cn}
  
\author{Cong Bai}
\orcid{0000-0002-6177-3862}
\affiliation{
  \institution{Zhejiang University of Technology}
  \department{College of Computer Science}
  \city{Hangzhou}
  \state{Zhejiang}
  \country{China}}
\email{congbai@zjut.edu.cn}

\author{Feng Zhang}
\orcid{0000-0003-4373-4058}
\affiliation{
  \institution{Fudan University}
  \department{Department of Atmospheric and Oceanic Sciences and Institutes of Atmospheric Sciences}
  \city{Shanghai}
  \country{China}}
% \affiliation{
%   \institution{Shanghai Key Laboratory of Ocean-land-atmosphere Boundary Dynamics and Climate Change}
%   \department{Key Laboratory of Polar Atmosphere-ocean-ice System for Weather and Climate of Ministry of Education}
%   \city{Shanghai}
%   \country{China}
% }
\email{fengzhang@fudan.edu.cn}

\renewcommand{\shortauthors}{Jiakang Shen et al.}

\begin{abstract}
Tropical cyclones (TC) are among the most destructive natural disasters, causing catastrophic damage to coastal regions through extreme winds, heavy rainfall, and storm surges. Timely monitoring of tropical cyclones is crucial for reducing loss of life and property, yet it is hindered by the computational inefficiency and high parameter counts of existing methods on resource-constrained edge devices. Current physics-guided models suffer from linear feature interactions that fail to capture high-order polynomial relationships between TC attributes, leading to inflated model sizes and hardware incompatibility. To overcome these challenges, this study introduces the Kolmogorov–Arnold Network-based Feature Interaction Framework (KAN-FIF), a lightweight multimodal architecture that integrates MLP and CNN layers with spline-parameterized KAN layers. For Maximum Sustained Wind (MSW)  prediction, experiments demonstrate that the KAN-FIF framework achieves a 94.8\% reduction in parameters (0.99MB vs 19MB) and 68.7\% faster inference per sample (2.3ms vs 7.35ms) compared to baseline model Phy-CoCo, while maintaining superior accuracy with 32.5\% lower MAE. The offline deployment experiment of the FY-4 series meteorological satellite processor on the Qingyun-1000 development board achieved a 14.41ms per-sample inference latency with the KAN-FIF framework, demonstrating promising feasibility for operational TC monitoring and extending deployability to edge-device AI applications. The code is released at https://github.com/Jinglin-Zhang/KAN-FIF.
\end{abstract}




\maketitle

\section{Introduction}
Tropical cyclones (TC), recognized as highly destructive meteorological phenomena, inflict severe damage upon coastal regions globally. Characterized by extreme wind velocities, intense precipitation, and elevated sea levels, these recurrent events primarily devastate infrastructure and communities through destructive sustained winds, extreme rainfall inducing catastrophic flooding, and devastating storm surges that overwhelm coastal defenses and cause widespread inundation. Consequently, accurate and timely monitoring of Tropical Cyclone (TC) intensity and the radius of peak winds is essential for effective disaster risk management, reliable warnings, urgent protective actions, and the guidance of evacuations.

Ground-centralized inference and edge-device inference are the two most popular TC prediction methods. The ground-centralized method involves multiple data transmission steps among satellite-to-ground and ground networks, resulting in high latency and computational costs. Edge-device inference faces stringent constraints in processing capability, memory, and operator compatibility. \cite{zhang2025s2dbft,zhang2012implementation,zhou2021novel} These limitations hinder the adoption of state-of-the-art MLP-based neural networks with a large number of CNN layers included, which are computationally intensive and require specialized hardware support.

Early TC estimation methods relied on satellite imagery combined with empirical algorithms such as the Dvorak technique\cite{Velden2006}. These approaches suffered from subjectivity and limited accuracy due to insufficient feature representation. With advances in deep learning, CNN-based models \cite{Wimmers2019, Zhang2021}  have dominated the estimation of TC attributes using spatial patterns in satellite data. However, most related works focus on the prediction of a single task, neglecting the inherent physical relationships between the TC attributes. Recent multitask learning (MTL) methods\cite{lee2020} attempt to share parameters among tasks, but they risk negative transfer and oversimplified feature interactions. Some works introduced physics-based constraints to model task correlations\cite{Zhou2023}, yet their linear feature concatenation fails to capture high-order polynomial relationships between variables while inflating parameter counts, further straining edge-device inference. Meanwhile, multiscale Xception networks with dual attention\cite{Ma2024} and YOLO-NAS architectures\cite{Nandal2025} achieve breakthroughs in fast satellite data processing by infrared-water vapor fusion. However, these deep learning approaches have advanced TC estimation accuracy at the cost of substantial computational demand.

Integrating multimodal auxiliary data with satellite imagery has emerged as a promising trend \cite{Huang2022,miao2021automated,li2021clothing,zhang2020ensemble}. However, related studies usually treat temporal features as static inputs rather than modeling sequential dependencies. Recurrent neural networks (RNN) and long-short-term memory networks (LSTM)\cite{kumar2023, Chen2021} have been applied to track prediction. The FuXi-based perturbation generator demonstrates improved TC track prediction by optimizing initial uncertainties \cite{Pu2025}. It operates independently of satellite data processing and requires extensive historical data for testing, creating notable constraints for field-deployed edge devices with limited memory capacity and demanding temporal requirements for timely tropical cyclone disaster monitoring.

\begin{figure}[t]
  \centering
  \includegraphics[width=\linewidth]{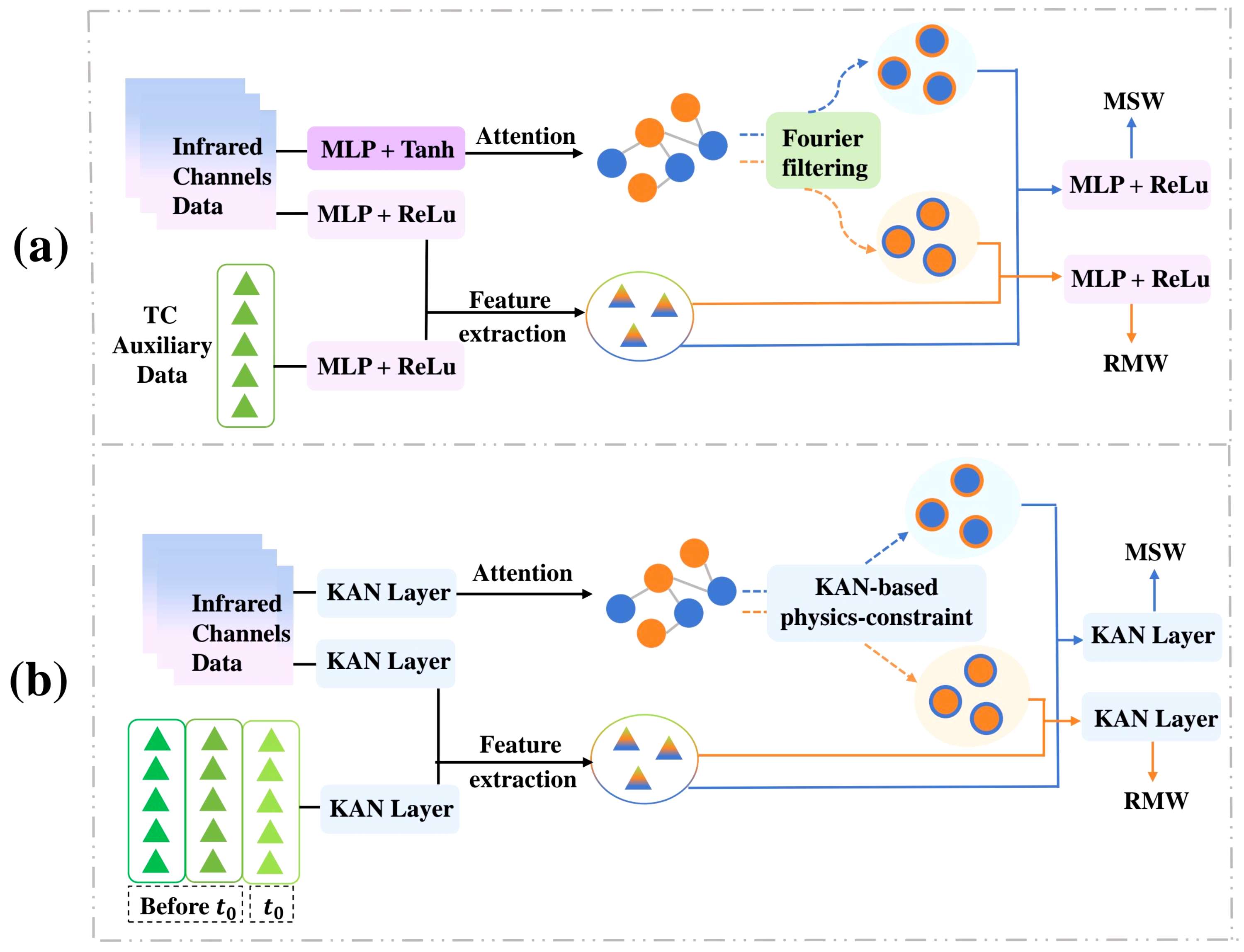} 
  \caption{The comparison of frameworks between the MLP-based and KAN-based models. (a) Conventional MLP-based model (Phycoco) (b) our KAN-based lightweight model
}
  \Description{The comparison of frameworks between the MLP-based and KAN-based models}
  \label{fig:intro} 
\end{figure}

These accumulated limitations pose three critical challenges for timely tropical cyclone prediction\cite{li2023fault,gao2020learning,zhu2022multi,feng20213d}: Firstly, current prediction methodologies predominantly rely on computationally intensive MLP/CNN architectures augmented with specialized detection heads and preprocessing filters, which substantially inflate parameter counts and necessitate high-bandwidth data transmission. Secondly, satellite data downlinking faces intrinsic constraints, such as asynchronous data accessibility due to confinement to orbital passes over specific regions, coupled with severe bandwidth limitations that restrict transmission rates. Thirdly, although edge-device inference offers potential solutions to downlink delays by enabling inter-satellite data transfer, processing, and inference entirely in-orbit with only minimal results relayed to ground stations, fundamental hardware barriers persist. These barriers encompass restricted computational capacity, insufficient memory resources, and framework-device compatibility issues, which collectively demand ultra-lightweight model architectures and preclude sophisticated multimodal preprocessing algorithms.

\begin{figure}[t]
  \centering
  \includegraphics[width=\linewidth]{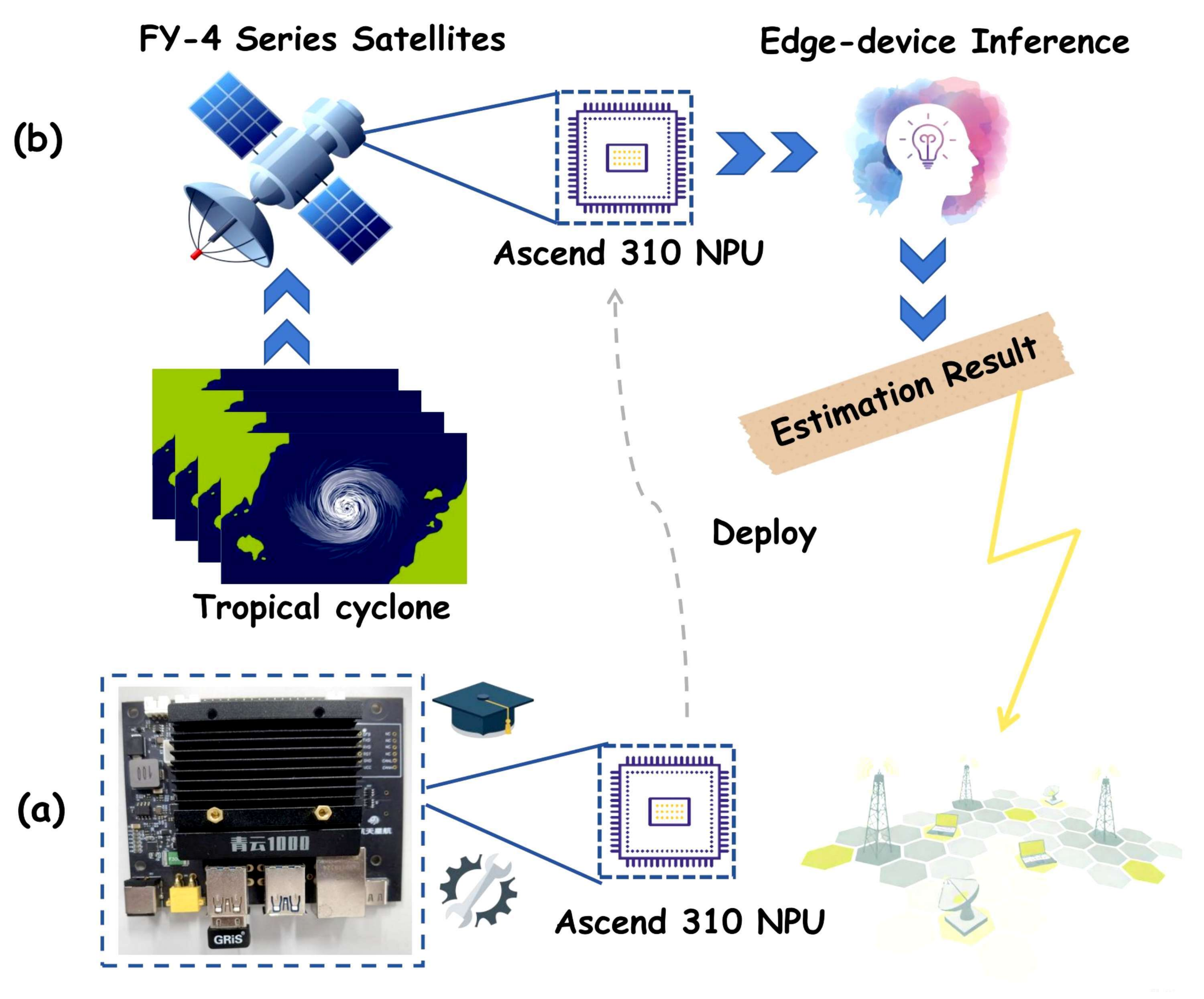} 
  \caption{The offline deployment verification process of the FY-4 series satellite processor on the Qingyun-1000 development board. (a) Deployment verification on Ascend 310 NPU (b) Edge-device inference process of tropical cyclone estimation on FY-4 series satellite}
  \Description{The offline deployment verification process of the FY-4 series satellite processor on the Qingyun-1000 board.}
  \label{fig:Deployment_process} 
\end{figure}

To address these challenges, the proposed \textbf{KAN-FIF} framework targets efficient and accurate tropical cyclone (TC) estimation on resource-constrained edge devices, with three core contributions:
\begin{itemize}
\item \textbf{Lightweight deployment via Kolmogorov-Arnold network (KAN) layer substitution.}
Traditional multilayer perceptrons (MLPs), convolutional neural networks (CNNs), and filtering operations in feature extraction, physical constraints, and decoding stages are replaced by computationally efficient KAN layers, significantly compressing model complexity while retaining superior accuracy, as shown in Figure \ref{fig:intro}.
\item \textbf{Physics-based fusion for cross-modal dependencies.} 
Sequential features and infrared imagery are fused through a hybrid encoder to capture nonlinear cross-modal couplings. Simultaneously, a differentiable physics-constraint module fits high-dimensional polynomial relationships between intensity and size by embedding domain-specific equations.
\item \textbf{On-Orbit Edge-device Inference Capability Validation.}
Experimental offline deployment of the FY-4 meteorological satellite processor on the Qingyun-1000 development board yielded a \textbf{14.41ms} latency per inference sample, with the verification process depicted in Figure \ref{fig:Deployment_process}. Through architectural refinements, edge-computing constraints were mitigated, enabling high-performance, timely tropical cyclone monitoring and demonstrating promising potential for operational deployment onboard satellites.
\end{itemize}
% 2 Related Work
\section{Related Work}
Despite advances in developing hardware-aware computational methods, persistent constraints continue to challenge multimodal data processing in tropical cyclone estimation. Consequently, advancing lightweight architectures must become a research priority to meet edge-device inference demands.
\subsection{Tropical Cyclone Estimation}
Tropical cyclone estimation has long been a focal point of extensive research, encompassing key attributes such as Maximum Sustained Wind (MSW), the radius of maximum winds (RMW), the spatial extent of impact, and associated precipitation volumes. Related models primarily utilize two data sources: infrared (IR) channel data from satellites, and auxiliary information comprising geographical metadata, seasonal context, and temporal characteristics. Although track prediction relies heavily on spatiotemporal sequences, the estimation of other attributes of TC usually depends on high-dimensional IR satellite imagery, like DeepMicroNet, TCIEnet, and TCICEnet\cite{Wimmers2019,Zhang2019,Zhang2021}. Recently, multi-task prediction frameworks have emerged that incorporate physical priors to constrain the output from multiple tasks\cite{Huang2023, Tian2022}. Some studies employ encoder-decoder fusion architectures or integrate frequency domain features through filtering operations\cite{chavas2017,xi2020,Yan2024}. However, these approaches usually rely on substantial computational resources and extensive parameters to capture non-linear relationships.
\subsection{Edge-device inference}
Edge-device inference offers a promising alternative to ground-centralized processing by mitigating limitations such as intermittent connectivity and restricted data transmission capacity. Early research targeted basic inference tasks, exemplified by CubeSat cloud segmentation studies \cite{Nagarajan2014}. Subsequent missions, exemplified by NASA's IPEX mission, achieving real-time classification \cite{Altinok2016}, revealed persistent computational constraints onboard. The rise of deep learning substantially advanced onboard capabilities, with recent missions validating high-accuracy cloud detection using optimized deep learning models \cite{Giuffrida2021}. Researchers subsequently explored efficient architectures through three pathways: First, lightweight models deployed on microcontrollers \cite{Park2020, Oktaviani2023}; Second, quantized networks achieving real-time performance on FPGAs \cite{Pitonak2022}; and Third, specialized designs such as row-wise stream processors \cite{Bahl2022}.

Separately in meteorological forecasting, lightweight models have optimized efficiency-accuracy tradeoffs. Tian et al. developed a multitask network termed TC-MTLNet with adaptive loss balancing for joint tropical cyclone intensity and size estimation, reducing errors while minimizing computational overhead \cite{Tian2022}. Similarly, Shang et al. proposed CDC-Net, which leverages channel dilation combined with feature copying to enable rapid satellite image classification under strict computational constraints \cite{Shang2023}. Building on these advances, our method integrates Kolmogorov-Arnold Network layers as substitutes for conventional CNN and MLP components. This design eliminates preprocessing filtration requirements while achieving dramatic parameter reduction, substantially enhancing feasibility for edge deployment.

\section{Problem Definition}
Our model targets predictions of two critical tropical cyclone attributes: Maximum Sustained Wind (MSW) and Radius of Maximum Wind (RMW), learning parametric mappings that:
\begin{equation}
\begin{array}{l}
f_{\text{msw}}:(X_{\text{seq}}, X_{\text{img}}) \rightarrow \widehat{Y}_{\text{msw}} \\

f_{\text{rmw}}:(X_{\text{seq}}, X_{\text{img}}) \rightarrow \widehat{Y}_{\text{rmw}}
\end{array}
\end{equation}

The TC estimation problem is formulated as:
\begin{equation}
\begin{split}
\theta^{*} = \underset{\theta}{\arg\max} \Bigg[ 
\alpha & \cdot L\left(f_{msw}(X_{seq}, X_{img}; \theta), Y_{msw}\right) \\ 
+ \beta & \cdot L\left(f_{rmw}(X_{seq}, X_{img}; \theta), Y_{rmw}\right) 
\Bigg]
\end{split}
\end{equation}
where $\theta$ represents all the learnable parameters.  $\alpha$ and $\beta$ represent the adjustable weights of the tasks. $L(\cdot)$ denotes the MAE loss function. $Y_{msw}$  and $Y_{rmw}$ represent ground truth values. $\widehat{Y}_{msw} \in [19,170]$ knots and $\widehat{Y}_{rmw} \in [5,200]$ nmi are denormalized through min-max scaling.
The input data consists of two modalities: 1) Temporal sequence data $X_{seq} \in \mathbb{R}^{T \times 5}$ containing the evolution features of the TC (latitude $x^{\text{lat}}$, longitude $x^{\text{lon}}$, the time since the TC was named $x^{t}$, previous category $x^{cat}$ and central pressure $x^{pre}$) over $T=3$ consecutive time steps, with a temporal resolution of 3 hours between steps. 2) Satellite image data $X_{img} \in \mathbb{R}^{8 \times H \times W}$ representing multichannel infrared observations with spatial dimensions $H=156$, $W=156$. The first four channels $X_{img}^{ch_{1-4}}$ correspond to observations from 3 hours before the target prediction time, while the last four channels $X_{img}^{ch_{5-8}}$ represent current-time imagery.

\begin{figure*}[t!]
  \centering
  \includegraphics[width=0.95\textwidth]{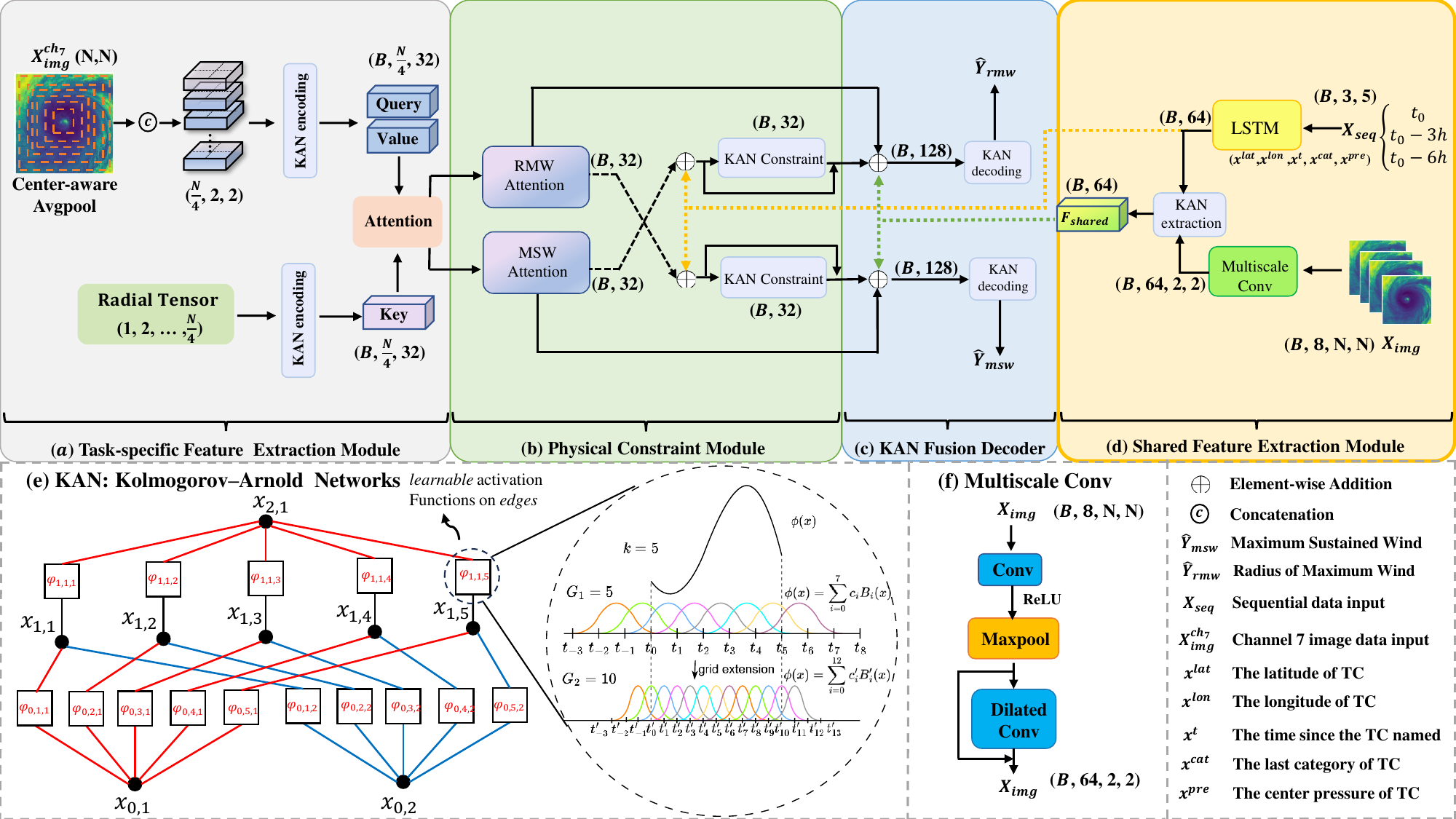}
  \caption{The architecture of the KAN-FIF learning framework: (a) The Task-specific Feature Extraction Module uses KAN layer and center-aware attention to extract the task features of MSW and RMW respectively  (b) The Physical Constraint Module is designed to conduct constraints among task-specific features ; (c) The KAN Fusion Decoder fuse the features from(a)(b)(d) and obtain the final output ; (d) The Shared Feature Extraction Module take the multi-channel image and the temporal sequence data as input and obtain the shared feature between TC tasks; (e) the architecture of KAN layers; (f)  the architecture of Multiscale Conv}
  \Description{The architecture of the KAN-FIF learning framework: (a) The Task-specific Feature Extraction Module uses KAN layer and center-aware attention to extract the task features of MSW and RMW respectively  (b) The Physical Constraint Module is designed to conduct constraints among task-specific features ; (c) The KAN Fusion Decoder fuse the features from(a)(b)(d) and obtain the final output ; (d) The Shared Feature Extraction Module take the multi-channel image and the temporal sequence data as input and obtain the shared feature between TC tasks; (e) the architecture of KAN layers; (f)  the architecture of Multiscale Conv}
  \label{fig:framework}
\end{figure*}

% 4 Methods
\section{Methods}
In this study, we integrate KAN layers in four critical aspects of our architecture, as shown in Figure~\ref{fig:framework}, in order to design a framework that takes into account both lightweight deployability and prediction accuracy:
a) \textbf{Shared Feature Extraction}: We employ KAN-LSTM for temporal feature extraction and KAN-CNN for spatial feature extraction from multi-spectral satellite imagery.
b) \textbf{Attention Encoding}: When computing center-aware spatial attention, we utilize KAN layers to encode both distance features and spatial patterns, replacing traditional linear+tanh encoding.
c) \textbf{Physical Constraints}: We directly implement inter-task physical constraints through KAN layers to fit polynomial relationships among task features, eliminating conventional convolution, MLP, Fourier transform, and Kalman filtering operations.
d) \textbf{Feature Fusion and Decoding}: We utilize the KAN layers to fuse and decode shared features, task-specific features, and physical constraint features to obtain the final prediction outputs.
Note that the full KAN-FIF architecture employs an LSTM layer followed by KAN projection for temporal features, while the deployment variant (Section~\ref{sec:deployment}) removes the LSTM due to NPU incompatibility. 
%4.1
\subsection{Kolmogorov-Arnold Networks}

Kolmogorov-Arnold networks (KANs) introduce a novel neural network framework inspired by the Kolmogorov-Arnold representation theorem \cite{Liu2024}. Unlike traditional Multi-Layer Perceptrons (MLPs) that employ fixed node activation functions, KANs utilize spline-approximated learnable activations parameterized on edges. This architectural innovation enables KANs to model complex nonlinear relationships with exceptional parameter efficiency, offering distinct advantages for high-dimensional data modeling. The foundation of KANs lies in the Kolmogorov-Arnold theorem, which asserts that any continuous multivariate function can be decomposed into a finite composition of univariate functions and additive operations. Mathematically, for a function $f:[0,1]^{n} \rightarrow \mathbb{R}$, this is expressed as:
\begin{equation}
f(x_1,\ldots,x_n) = \sum_{q=1}^{2n+1} \varphi_q \left( \sum_{p=1}^n \varphi_{q,p}(x_p) \right)
\end{equation}
where $\varphi_{q,p}$ and $\varphi_q$ are univariate functions. 

KAN layers fundamentally replace linear weight matrices with spline-parameterized edges. This architectural shift enables adaptive modeling of high-order polynomial relationships, where MLP and CNN architectures struggle due to fixed activation functions constraining expressiveness. By leveraging locally adaptive polynomial segments through edge-based parameterization, KAN layers efficiently capture intricate nonlinearities without requiring excessive network depth. Crucially, as KAN layers directly optimize inter-layer function parameters, their outputs form strictly expressible composite polynomials, facilitating significantly higher interpretability than MLPs through explicit exploration of variable relationships. Each edge in a KAN layer, as shown in Figure~\ref{fig:framework}(e), represents a univariate function $\varphi_{q,p}$ parameterized as:
\begin{equation}
\varphi_{q,p}(x) = w \cdot \left( \text{silu}(x) + \text{spline}(x) \right)
\end{equation}
where $\text{silu}(x) = \frac{x}{1 + e^{-x}}$ is a self-gating activation function
and $\text{spline}(x)$ is a piecewise polynomial curve optimized via gradient descent.

In subsequent implementations, the KAN layer can be treated as a linear layer, employing fixed polynomial basis functions with grid\_size=5 and spline\_order=3. Here, grid\_size specifies the number of intervals used to partition the input domain for the spline approximation, where each interval corresponds to a locally defined polynomial segment. The spline\_order parameter controls the degree of the polynomial used in each interval, with spline\_order=3 corresponding to cubic splines.

%4.2
\subsection{KAN-based Shared Feature Extraction}

The Shared Feature Extraction Module implements parallel temporal-spatial feature extraction through dual pathways. \textbf{KAN-LSTM} processes sequential TC evolution data and \textbf{KAN-CNN} extracts multi-scale spatial patterns from infrared imagery.

\subsubsection{Temporal Feature Extraction with KAN-LSTM}
For temporal input $X_{\text{seq}} \in \mathbb{R}^{B \times T \times 5}$ containing $B$ samples of 3-step TC evolution features, we first employ an LSTM layer to capture temporal dependencies, and then a KAN projection extracts the features:
\begin{equation}
F_{\text{seq}} = \text{KAN}_{\text{Linear}}^{64 \rightarrow 32}\left(F_{\text{LSTM}}\right) \in \mathbb{R}^{B \times 32}
\end{equation}

\subsubsection{Spatial Feature Extraction with KAN-CNN}
For satellite imagery $X_{img} \in \mathbb{R}^{B \times 8 \times H \times W}$, we design a Multi-Scale ConvBlock with residual enhancement:
\begin{equation}
\begin{aligned}
F_{\text{conv}}^{(1)} &= \text{ReLU}\left(\text{Conv2D}_{8 \rightarrow 16}^{5 \times 5}\left(X_{img}\right)\right) \\
F_{\text{conv}}^{(2)} &= \text{MaxPool}\left(\text{ReLU}\left(\text{Conv2D}_{16 \rightarrow 32}^{3 \times 3}\left(F_{\text{conv}}^{(1)}\right)\right)\right) \\
F_{\text{res}} &= \text{Conv2D}_{32 \rightarrow 64}^{1 \times 1}\left(F_{\text{conv}}^{(2)}\right) \quad \text{(Residual path)} \\
F_{\text{multiscale}} &= \text{concat}\left[F_{\text{res}}, \sum_{d=1}^{3} \text{DilatedConv}_{32 \rightarrow 32}^{3 \times 3}\left(F_{\text{conv}}^{(2)}\right)\right]
\end{aligned}
\end{equation}
The compressed spatial features are obtained as follows:
\begin{equation}
F_{\text{img}} = \text{KAN}_{\text{Linear}}^{256 \rightarrow 32}\left(\text{Flatten}\left(F_{\text{multiscale}}\right)\right) \in \mathbb{R}^{B \times 32}
\end{equation}

\subsubsection{Final Shared Representation}
The final shared representation combines temporal and spatial features:
\begin{equation}
F_{\text{shared}} = \text{concat}\left[F_{\text{seq}}, F_{img}\right] \in \mathbb{R}^{B \times 64}
\end{equation}

%4.3
\subsection{KAN-based Task-Specific Feature Extraction}

Traditional attention mechanisms for TC estimation often employ linear layers with static activation functions (ReLU or tanh) to encode spatial distances and cloud patterns. However, these methods struggle to model high-order polynomial relationships between annular cloud features and TC attributes. In our KAN-Attention, we replace linear layers with KAN layers in spatial distance encoding and content-aware feature projection(Algorithm~\ref{alg:kan-attention}).

We selected channel 7 (10.4 $\mu m$ infrared band from Himawari-8) as spatial input because it optimally captures cold cloud tops in TC eye-wall regions, where lower brightness temperatures correlate strongly with convective intensity and MSW/RMW values\cite{Yan2024}. Our preprocessing employs adaptive annular pooling across 39 concentric rings to extract hierarchical cloud features, generating content-aware queries (Q) and values (V). Radial distances are encoded through a KAN layer to produce position-sensitive keys (K). 

\begin{algorithm}[tb]
\caption{KAN-Attention}
\label{alg:kan-attention}
\begin{algorithmic}[1]
\REQUIRE $F_{\text{seq}} \in \mathbb{R}^{B\times32}$: Temporal features \\
\REQUIRE $X_{\text{img}}^{ch_7} \in \mathbb{R}^{B\times1\times H\times W}$: Channel 7 image \\ \textbf{Output} $A_{\text{task}} \in \mathbb{R}^{B\times32}$
\STATE \textbf{// Center-aware Avgpool}
\STATE Initialize $r\_center \gets 77$
\FOR{$i \in \{0,\ldots,38\}$}
    \STATE $L \gets r\_center - 2i$, $R \gets r\_center + 2i$
    \STATE $R_i \gets X_{\text{img}}[:,:,L:R,L:R]$
    \STATE $P_i \gets \text{AdaptiveAvgPool}(R_i,(2,2))$
    \STATE $P \gets \text{Concat}(P, \text{Flatten}(P_i))$
\ENDFOR
\STATE Initialize $G \gets \text{linspace}(0,1,39)$
\STATE \textbf{// KAN-based encoding}
\STATE $K \gets \text{KAN\_Linear}(G)$ \COMMENT{Spatial distance encoding}
\STATE $Q,V \gets \text{Split}(\text{KAN\_Linear}(P),2)$ \COMMENT{Content encoding}
\STATE \textbf{// Multi-head attention}
\STATE $Q_h, K_h, V_h \gets \text{SplitHeads}(Q, K, V, \text{num\_heads}=4)$
\STATE $\text{Attn} \gets \text{Softmax}\left(Q_h @ K_h^T / \sqrt{d}\right)$
\STATE $C_h \gets \text{Attn} @ V_h$, $C \gets \text{MergeHeads}(C_h)$
\STATE \textbf{// Temporal fusion}
\STATE $A_{\text{task}} \gets \text{KAN\_Linear}(\text{Concat}(C_{avg}, F_{\text{seq}}))$
\end{algorithmic}
\end{algorithm}

\subsection{Physics-Guided Constraint Modeling}
\label{subsec:physics_constraint}

To establish meteorologically consistent relationships between MSW and RMW predictions, we design bidirectional residual connections governed by KAN layers. Let $A_{msw}\in\mathbb{R}^{B \times 32}$ and $A_{rmw}\in\mathbb{R}^{B \times 32}$ be the task-specific features. The physics-guided interaction is formulated as follows.
\begin{equation}
\begin{aligned}
\quad & \Gamma_{msw \rightarrow rmw}(A_{msw}) = A_{rmw} + K_{msw2rmw}(A_{msw}) \\
\quad & \Gamma_{rmw \rightarrow msw}(A_{rmw}) = A_{msw} + K_{rmw2msw}(A_{rmw})
\end{aligned}
\end{equation}
where $K(\cdot)$ represents a KAN layer that implements a dimensional mapping. $K_{msw2rmw}$ and $K_{rmw2msw}$ employ independent spline bases to model distinct physical mechanisms: $K_{msw2rmw}$ learns wind-driven radius expansion or contraction patterns and $K_{rmw2msw}$ captures radius-modulated wind intensification. This formulation preserves the primary task features through residual connections while injecting physically plausible interactions between TC attributes.

\subsection{Multimodal Feature Fusion}
\label{subsec:multimodal_fusion}
Final predictions fuse three categories of features:
\begin{enumerate}
  \item Task-specific features ($A_{msw}$, $A_{rmw}$)
  \item Physics-constrained features ($\Gamma_{msw\rightarrow rmw}$, $\Gamma_{rmw\rightarrow msw}$)
  \item Shared spatiotemporal embeddings $F_{\text{shared}}$
\end{enumerate}

The fusion and decoding process is defined as follows:
\begin{equation}
\begin{aligned}
\bar{Y}_{msw} &= D_{msw}\left(\text{cat}\left[A_{msw}, \Gamma_{rmw \rightarrow msw}, F_{\text{shared}}\right]\right) \\
\bar{Y}_{rmw} &= D_{rmw}\left(\text{cat}\left[A_{rmw}, \Gamma_{msw \rightarrow rmw}, F_{\text{shared}}\right]\right)
\end{aligned}
\end{equation}
where $D(\cdot)$ denotes a KAN decoder which can be configured as follows.
\begin{equation}
D_{\text{task}}(x) = \text{KAN}_{\text{Linear}}^{128 \rightarrow 1}(x)
\end{equation}

% 5 deployment
\section{Deployment Preparation for TC Estimation}
\label{sec:deployment}

\subsection{Hardware and Software Environment}
\label{subsec:hardware_env}
The proposed KAN-FIF framework was experimentally deployed on the Qingyun-1000 development board for the FY-4 meteorological satellite processor, and three edge-deployment constraints were overcome: limited operator compatibility for neural network layers, hardware acceleration dependencies requiring static computation graphs, and memory limitations necessitating model compression to prevent runtime failures. 
\begin{table}[ht]
\centering
\caption{Hardware and Software Deployment Specifications}
\label{tab:deployment-specs}
\begin{tabular}{@{}l@{\hspace{20pt}}l@{}}
\toprule
\textbf{Category} & \textbf{Specification} \\
\midrule
\textbf{Hardware Platform} & Qingyun-1000 board\\
\textbf{Acceleration Module} & Atlas 200I A2 \\
\textbf{Processor} & Huawei Ascend 310 NPU \\
\textbf{Memory} & 8GB LPDDR4 \\
\textbf{Compute Capacity} & 22 TOPS @ INT8/ \\& 11 TOPS @ FP16 \\
\textbf{Power Budget} & $<$10W \\
\midrule
\textbf{Acceleration Stack} & CANN 6.0.1 \\
\textbf{Environment} & Python 3.7 \\
\bottomrule
\end{tabular}
\end{table}
The edge-device deployment leverages Huawei's AscendCL framework to replace standard PyTorch inference with a static execution paradigm. Key adaptations involve initializing inference sessions via InferSession for loading precompiled OM models, enforcing FP16 tensor precision to ensure NPU compatibility, and implementing asynchronous I/O queues to minimize latency. Unlike GPU-based inference relying on dynamic computation graphs, this deployment pre-allocates fixed-size input buffers for satellite imagery while eliminating runtime branching operations.
\subsection{Model Conversion Workflow}
The original KAN-LSTM hybrid architecture underwent redesign to eliminate NPU-incompatible LSTM operators. In its deployment version, flattened temporal sequences are processed directly through an expanded KAN network, bypassing recurrent computations entirely.

To ensure compliance with the NPU's static computation graph requirements, all conditional branching operations were removed to establish a fixed execution path. Key hyperparameters—including the spline grid\_size (fixed at 5) and spline\_order (fixed at 3)—were hardcoded, while B-spline basis functions underwent precomputation during model initialization. Adaptive pooling layers were replaced by fixed-stride average pooling, preventing dynamic output shapes. Collectively, these modifications guarantee deterministic memory allocation and operator sequencing during inference.

For addressing the ATC compiler's 63-kernel-size limit on pooling operations, the deployment version employs a two-stage pooling strategy.

While our offline validation demonstrates promising performance on hardware identical to the FY-4 satellite processor, actual on-orbit deployment may face additional challenges including data pipeline integration, radiation hardening, power management, and command-control systems that require further engineering validation.

% 实验部分
\section{Experiments}

\subsection{Experimental Setup}
\label{subsec:exp_setup}

\subsubsection{Dataset and Preprocessing}
The model is evaluated using the Tropical Cyclone Multi-Modal Dataset (TCMM), which shares data sources with the baseline study \cite{Yan2024,huang2025} but incorporates temporal sequencing. This dataset integrates infrared brightness temperature from Himawari-8 satellite channels 7, 8, 13, 15 \cite{Bessho2016} and tropical cyclone track records from the IBTrACS repository \cite{Knapp2010,Gahtan2024}, covering the period 2015-2022. Each sample comprises an 8-channel $156\times156$ infrared image normalized to $[0,1]$, paired with five temporal features: (a) center position (latitude/longitude), (b) hours since cyclone naming, (c) prior storm category (0-5 scale with -1 for unnamed systems), and (d) central pressure (hPa). Samples are structured as 3-hour interval sequences ($t_{0}-6$, $t_{0}-3$, $t_0$) to capture short-term evolution, with this 6-hour lookback window enabling operational forecasts 6 hours post-naming—critical for timely disaster prevention. To enhance dataset robustness, we applied  $90^\circ$ clockwise, $90^\circ$ counterclockwise, and $180^\circ$ rotational augmentation to the original images. Augmented data are stratified by cyclone name and rotation type to preserve spatiotemporal consistency during training and evaluation.
\subsubsection{Implementation Details}
Model training is conducted on an NVIDIA RTX 4090 GPU using PyTorch, employing Stochastic Gradient Descent (SGD) with an initial learning rate of 0.001 and batch size of 128. Training proceeds for a maximum of 200 epochs, with early stopping typically triggered at approximately epoch 20 upon validation loss plateauing. To mitigate overfitting and accelerate convergence, we implement learning rate scheduling (reducing by a factor of 0.5 after 5 epochs of validation loss stagnation) and early stopping (activated after 10 epochs without improvement). Adhering to a strict temporal holdout strategy to prevent information leakage, the data partition comprises: training set (2015-2020, 46,285 samples), validation and test set (2021-2022, 1,140 and 1,158 samples respectively). 
\subsubsection{Evaluation Metrics}
In alignment with standard tropical cyclone estimation protocols, we evaluate two key metrics for both Maximum Sustained Wind (MSW) and Radius of Maximum Winds (RMW): the mean absolute error  $L_{\text{MAE}}(\hat{Y}, Y) = \frac{1}{N}\sum_{i=1}^{N} |\hat{Y}_i - Y_i|$ and root mean square error $L_{\text{RMSE}}(\hat{Y}, Y) = \sqrt{\frac{1}{N}\sum_{i=1}^{N} (\hat{Y}_i - Y_i)^2}$. All metrics are computed on denormalized predictions using operational units—MSW in knots (range: [19, 170 kt]) and RMW in nautical miles (range: [5, 200 nmi])—ensuring direct comparability with established literature and operational forecasting systems.
\subsection{Comparison with State-of-the-Art Methods}
\label{subsec:sota_comparison}

\begin{figure*}[t!]
  \centering
  \includegraphics[width=1\textwidth]{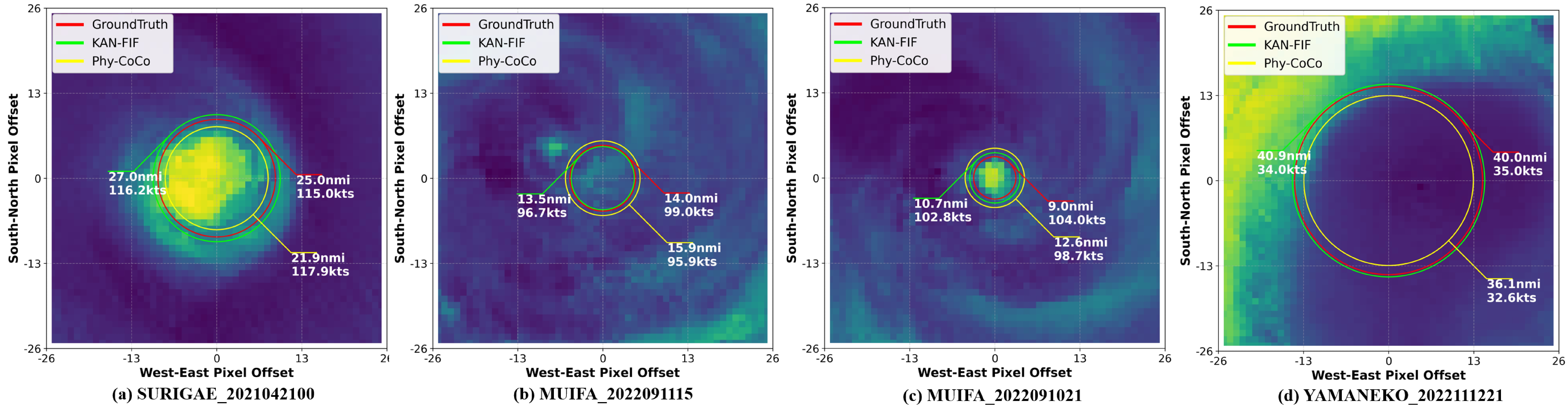}
  \caption{Visual comparison between KAN-FIF and Phy-CoCo models under different tropical cyclone structures}
  \Description{Visual comparison between KAN-FIF and Phy-CoCo models under different tropical cyclone structures}
  \label{fig:visualise}
\end{figure*}

We compare KAN-FIF with seven established tropical cyclone estimation methods across four accuracy metrics, as shown in Table~\ref{tab:sota_comparison}. The complete KAN-FIF model achieves a 32.5\% reduction in MSW MAE (3.21~kt vs.~4.76~kt) and a 31.9\% RMSE reduction (4.31~kt vs.~6.33~kt) compared to the state-of-the-art multi-task model Phy-CoCo. RMW prediction shows consistent improvement (MAE: 8.83~nmi vs.~8.89~nmi), validating KAN-FIF's ability to model cross-task physical relationships without negative transfer. Furthermore, while DeepMicroNet, TCIEnet, TCICEnet, and the Xception model also demonstrate competitive accuracy, these are single-task models focused solely on intensity prediction. It is evident that even when compared to these specialized models, KAN-FIF achieves superior predictive performance.
\begin{table}[ht]
\centering
\footnotesize 
\caption{Performance comparison of state-of-the-art TC estimation methods}
\label{tab:sota_comparison}
\begin{tabular}{@{}l@{\hspace{15pt}}c@{\hspace{15pt}}c@{\hspace{15pt}}c@{\hspace{15pt}}c@{}} 
\toprule
\textbf{Model}  & \multicolumn{2}{c}{\textbf{MSW}} & \multicolumn{2}{c}{\textbf{RMW}} \\
\cmidrule(lr){2-3} \cmidrule(l){4-5} % 添加分组横线
&   \textbf{MAE} & \textbf{RMSE} & \textbf{MAE} & \textbf{RMSE} \\
\midrule
TC-MTLNet\cite{Tian2022}  & 9.99 & 13.77 & 11.03 & 14.53 \\
DeepCNet\cite{Zhuo2021} & 6.84 & 9.25 & 11.21 & 15.10 \\
DeepMicroNet\cite{Wimmers2019} & 3.94 & 5.47 & \multicolumn{2}{c}{(single-task)} \\
TCIEnet\cite{Zhang2019} & 3.61 & 4.93 & \multicolumn{2}{c}{(single-task)} \\
TCICEnet\cite{Zhang2021}  & 3.47 & 4.75 & \multicolumn{2}{c}{(single-task)} \\
Xception\cite{Ma2024}  & 3.88 & 4.50 & \multicolumn{2}{c}{(single-task)} \\
Phy-CoCo\cite{Yan2024} & 4.76 & 6.33 & 8.89 & 12.24 \\
\midrule
\textbf{KAN-FIF}  & \textbf{3.21} & \textbf{4.31} & \textbf{8.83} & \textbf{11.66} \\
\bottomrule
\end{tabular}
\end{table}
Further Results demonstrate significant improvements in computational efficiency and estimation accuracy (Table~\ref{tab:size_and_time}), where KAN-FIF reduces parameter count by 94.8\% versus Phy-CoCo (from 19MB to 0.99MB) and decreases per-sample inference time by 68.7\% (7.35ms to 2.3ms), enabling lightweight edge-device deployment.
\begin{table}[ht]
\centering
\footnotesize 
\caption{Comparison of model size and inference time with multi-task model}
\label{tab:size_and_time}
\begin{tabular}{@{}c@{\hspace{4pt}}c@{\hspace{4pt}}c@{\hspace{4pt}}c@{\hspace{4pt}}c@{}} 
\toprule
\textbf{Model} &TC-MTLNet\cite{Tian2022} & DeepCNet\cite{Zhuo2021}& Phy-CoCo\cite{Yan2024} & \textbf{KAN-FIF} \\
\midrule
\textbf{Size(M)} &170 & 86 &  19&  \textbf{0.99}\\
\textbf{Infer Time(ms)}&10.17 & 8.91 & 7.35& \textbf{2.3}\\
\bottomrule
\end{tabular}
\end{table}
To comprehensively evaluate model performance, we selected four representative tropical cyclone structures from the test set for quantitative assessment. Figure~\ref{fig:visualise} provides visual comparisons between KAN-FIF predictions and the baseline Phy-CoCo model, where circle sizes scale proportionally with RMW values and MSW magnitudes are annotated along corresponding guidelines.
\subsection{Deployment metrics}
\label{subsec:Deployment_metrics}
Prior to deployment experiments, we executed operator compatibility refactoring, constructed static computation graphs, and performed offline model conversion (as detailed in Section~\ref{sec:deployment}) to produce the deployable KAN-FIF variant. Following these architectural simplifications, predictive accuracy exhibited marginal degradation within expected tolerances (Table~\ref{tab:deployment}, deploy(GPU)). Although static computation graph enforcement and LSTM removal contribute to this marginal accuracy loss, the model largely preserved baseline precision, demonstrating the robustness of KAN layers under operator compatibility constraints.

Leveraging the Ascend 310 NPU processor onboard FY-4 series meteorological satellites, we conducted offline deployment experiments on the Qingyun-1000 development board using multispectral remote sensing data and temporal auxiliary inputs. The edge-device deployment achieved a per-sample inference latency of 14.41ms, validating the promising potential for operational tropical cyclone monitoring.
\begin{table}[ht]
\centering
\footnotesize
\caption{Metrics of deployment}
\label{tab:deployment}
\begin{tabular}{@{}l@{\hspace{6pt}}c@{\hspace{6pt}}c@{\hspace{6pt}}c@{\hspace{6pt}}c@{\hspace{6pt}}c@{\hspace{6pt}}c@{}} 
\toprule
\textbf{Model} & \textbf{Size} & \textbf{Infer Time} & \multicolumn{2}{c}{\textbf{MSW}} & \multicolumn{2}{c}{\textbf{RMW}} \\
\cmidrule(lr){4-5} \cmidrule(l){6-7} 
& \textbf{(M)} & \textbf{(ms)} & \textbf{MAE} & \textbf{RMSE} & \textbf{MAE} & \textbf{RMSE} \\
\midrule
deploy(GPU) & 0.92 & 2.28 & 3.63 & 4.93 & 8.95 & 11.77 \\
deploy(AscendNPU) & 0.92 & 14.41 & 6.66 & 9.78 & 9.37 & 12.22 \\
\midrule
\textbf{KAN-FIF} & \textbf{0.99} & \textbf{2.30} & \textbf{3.21} & \textbf{4.31} & \textbf{8.83} & \textbf{11.66} \\
\bottomrule
\end{tabular}
\end{table}
\subsection{Expanded Results Analysis on Compressed Model}
Further demonstrating the scalability and deployment potential of KAN-FIF, experiments with substantially reduced hidden units achieved a 77.8\% parameter reduction (from 0.99M to 0.22M) while maintaining competitive performance (Table~\ref{tab:expanded_results} KAN-FIF-s). The resulting model exhibits minimal accuracy degradation, evidenced by marginal increases in MSW MAE (3.21 to 3.39 kt), RMW MAE (8.83 to 8.98 nmi), MSW RMSE (4.31 to 4.49 kt), and RMW RMSE (11.66 to 11.79 nmi). Crucially, this parameter efficiency reduces memory footprint while reducing on-board inference energy consumption.
\begin{table}[ht]
\centering
\footnotesize 
\caption{Performance comparison with compressed model}
\label{tab:expanded_results}
\begin{tabular}{@{}l@{\hspace{8pt}}c@{\hspace{8pt}}c@{\hspace{8pt}}c@{\hspace{8pt}}c@{\hspace{8pt}}c@{\hspace{8pt}}c@{}}
\toprule
\textbf{Model} & \textbf{Size} & \textbf{Infer Time} & \multicolumn{2}{c}{\textbf{MSW}} & \multicolumn{2}{c}{\textbf{RMW}} \\
\cmidrule(lr){4-5} \cmidrule(l){6-7} % 添加分组横线
& \textbf{(M)} & \textbf{(ms)} & \textbf{MAE} & \textbf{RMSE} & \textbf{MAE} & \textbf{RMSE} \\
\midrule
\textbf{KAN-FIF} & \textbf{0.99} & \textbf{2.30} & \textbf{3.21} & \textbf{4.31} & \textbf{8.83} & \textbf{11.66} \\
KAN-FIF-s & 0.22 & 2.30 & 3.39 & 4.49 & 8.98 & 11.79 \\
\bottomrule
\end{tabular}
\end{table}

\subsection{Ablation study}
\subsubsection{Ablation Analysis of Temporal Feature Integration}
To underscore the critical role of temporal feature organization in KAN-FIF, we first analyze the performance gap between KAN-FIF and the deployment variant, wherein the LSTM layer is substituted with KAN layers, resulting in marginal accuracy degradation (MSW MAE: +13.1\%, RMW MAE: +1.4\%). Conversely, when temporal features are removed and single-time-step inputs encoded via linear layers with tanh activation, severe performance deterioration occurs: MSW MAE increases by 32.08\% and RMSE by 32.01\%, while RMW MAE rises 7.70\% with RMSE increasing 8.58\%. As shown in Table~\ref{tab:temporal_ablation}, the stark contrast demonstrates that while edge deployment necessitates LSTM removal, preserving temporal sequencing through KAN projections largely maintains accuracy, whereas abandoning temporal context catastrophically compromises prediction accuracy.
\begin{table}[ht]
\centering
\footnotesize
\caption{Ablation study on temporal feature integration}
\label{tab:temporal_ablation}
\begin{tabular}{@{}c@{\hspace{10pt}}c@{\hspace{10pt}}c@{\hspace{10pt}}c@{\hspace{10pt}}c@{\hspace{10pt}}c@{}}
\toprule
\multicolumn{2}{c}{\textbf{Methods}} & \multicolumn{2}{c}{\textbf{MSW}} & \multicolumn{2}{c}{\textbf{RMW}} \\
\cmidrule(lr){1-2}\cmidrule(lr){3-4} \cmidrule(l){5-6} % 添加分组横线
\textbf{LSTM}&\textbf{Seq feature}& \textbf{MAE} & \textbf{RMSE} & \textbf{MAE} & \textbf{RMSE} \\ 
\midrule
 &  & 4.24 & 5.69 & 9.51 & 12.66 \\
 & \checkmark & 3.63 & 4.93 & 8.95 & 11.77 \\ 
\midrule
\multicolumn{2}{c}{\textbf{KAN-FIF}} & \textbf{3.21} & \textbf{4.31} & \textbf{8.83} & \textbf{11.66}\\ 
\bottomrule
\end{tabular}
\end{table}

\subsubsection{Hyperparameter Sensitivity Analysis}
Ablation studies on KAN hyperparameters revealed moderate sensitivity to \texttt{spline\_order} but limited sensitivity to \texttt{grid\_size}. The fixed configuration (grid\_size=5, spline\_order=3) was selected based on this analysis to balance accuracy and deployment constraints.

\subsubsection{Ablation Study on Multi-Stage KAN Integration}
This ablation study establishes the indispensable role of systematically integrating KAN layers across four architectural stages: shared feature extraction, attention-based encoding, physics-constrained modeling, and prediction decoding. By implementing KAN layers at each stage and benchmarking against the full KAN-FIF model (Table~\ref{tab:stage_ablation}), we quantify their contribution to accuracy gains through KAN layers. 

Five ablated variants were evaluated: 
 (a) Full substitution of KAN layers with MLPs (bilinear layers + ReLU/tanh activations) or Fourier transform modules 
 (b) Replacement of KAN layers in the Shared Feature Extraction Module (KAN-LSTM/KAN-CNN) with linear-ReLU blocks 
 (c) Substitution of KAN-Attention encoding with linear-tanh blocks 
 (d) Exchange of the KAN-based Physics-Constraint Module for the Fourier Transform Module from Phy-CoCo~\cite{Yan2024} 
 (e) Application of linear layers instead of KAN layers for final prediction decoding

\begin{table}[ht]
\centering
\footnotesize
\caption{Ablation Study on Multi-Stage KAN Integration}
\label{tab:stage_ablation}
\begin{tabular}{@{}c@{\hspace{3pt}}c@{\hspace{3pt}}c@{\hspace{3pt}}c@{\hspace{3pt}}c@{\hspace{3pt}}c@{\hspace{3pt}}c@{\hspace{3pt}}c@{}} 
\toprule
\multicolumn{4}{c}{\textbf{KAN Ablation Stage}} & \multicolumn{2}{c}{\textbf{MSW}} & \multicolumn{2}{c}{\textbf{RMW}} \\
\cmidrule(lr){1-4}\cmidrule(lr){5-6} \cmidrule(l){7-8}
 \textbf{Extract} & \textbf{Attention} & \textbf{Constraint} & \textbf{Decoder} & \textbf{MAE} & \textbf{RMSE} & \textbf{MAE} & \textbf{RMSE} \\ 
\midrule
&  & & & 3.63 & 4.85 & 9.96 & 13.24 \\ 
 & \checkmark& \checkmark& \checkmark& 3.46 & 4.62 & 9.15 & 11.94 \\ 
 \checkmark & &\checkmark & \checkmark & 3.41 & 4.54 & 8.84 & 11.87 \\ 
\checkmark &\checkmark & & \checkmark& 3.76 & 4.94 & 8.93 & 11.99 \\ 
\checkmark &\checkmark &\checkmark & & 3.53 & 4.66 & 9.12 & 11.69 \\ 
\midrule
 \multicolumn{4}{c}{\textbf{KAN-FIF}} & \textbf{3.21} & \textbf{4.31} & \textbf{8.83} & \textbf{11.66} \\
\bottomrule
\end{tabular}
\end{table}

The full KAN-FIF model achieves optimal performance in all metrics. The greatest degradation occurs when removing KAN layers from the physics-guided constraint stage(MSW MAE: +17.1\%), validating KANs' superiority in modeling high-order polynomial wind-radius relationships. While removing KAN layers from the attention-based encoding stage exhibits minimal degradation of RMW (+0.11\%), its MSW MAE increases by 6.2\%. The marginal impact on RMW estimation may suggest that the radius prediction relies more on shared spatial features than on task-specific ones. Removing KAN layers from the shared feature extraction stage degrades both tasks (MSW MAE: +7.8\%, RMW MAE: +3.6\%).

The cumulative degradation when removing KAN layers from all four stages (+13.1\% MSW MAE) reflects nonlinear interactions between ablation stages rather than simple additive effects. In particular, full-stage removal induces less severe degradation than isolated removal in the physics-guided constraint stage (which degrades MSW MAE by 17.1\%). This counterintuitive result suggests that MLP-based substitutions in non-critical modules partially compensate for errors introduced by removing physics-constrained KAN layers, masking the full impact of individual substitutions. However, such compensation is task-specific and unstable: RMW MAE degradation (+12.8\%) disproportionately exceeds isolated ablation impacts, highlighting the role of KAN layers in harmonizing multitask predictions.

\section{Conclusion}

In this study, we propose a resource-efficient framework KAN-FIF for tropical cyclone estimation of intensity and size, which integrates MLP, CNN, and filtering operations with spline-parameterized KAN layers to resolve computational and deployment constraints. The framework achieves state-of-the-art accuracy with a 94.8\% parameter reduction compared to physics-guided baselines, leveraging the mathematical foundation of KAN layers in multivariate function decomposition for adaptive high-order nonlinear modeling. Offline deployment on the Qingyun-1000 development board equipped with the Ascend 310 NPU, identical to the FY-4 satellite processor, achieved a 14.41ms per-sample inference latency, validating hardware readiness for orbital deployment through static computation graph optimization and memory-aware compression, establishing a scalable edge-device inference paradigm for high-precision geophysical modeling, offering significant societal value for disaster monitoring in infrastructure-limited regions. Future work will expand multimodal spatiotemporal datasets and optimize dynamic KAN configurations for meteorological satellite deployment scenarios.

\begin{acks}
This work was supported in part by the National Key R\&D Program of China under Grant 2022YFB3206900, Key R\&D Program of Shandong Province of China under Grant 2023CXGC010112, the joint funds of the National Natural Science Foundation of China under Grant U24A20221, Distinguished Young Scholar of Shandong Province under Grant ZR2023JQ025, Taishan Scholars Program under Grant tstp20250708, Major Basic Research Projects of Shandong Province under Grant ZR2022ZD32.
\end{acks}
\bibliographystyle{ACM-Reference-Format}
\balance
\bibliography{reference}

@article{velden2006,
  title={The Dvorak tropical cyclone intensity estimation technique: A satellite-based method that has endured for over 30 years},
  author={Velden, Christopher and Harper, Bruce and Wells, Frank and Beven, John L and Zehr, Ray and Olander, Timothy and Mayfield, Max and Guard, Charles “CHIP” and Lander, Mark and Edson, Roger and others},
  journal={Bulletin of the American Meteorological Society},
  volume={87},
  number={9},
  pages={1195--1210},
  year={2006},
  publisher={American Meteorological Society}
}

@article{lee2020,
  title={Multi-task learning based tropical cyclone intensity monitoring and forecasting through fusion of geostationary satellite data and numerical forecasting model output},
  author={Lee, Juhyun and Yoo, Cheolhee and Im, Jungho and Shin, Yeji and Cho, Dongjin},
  journal={Korean journal of remote sensing},
  volume={36},
  number={5\_3},
  pages={1037--1051},
  year={2020},
  publisher={Korean Society of Remote Sensing}
}

@article{Tian2022,
  title={A lightweight multitask learning model with adaptive loss balance for tropical cyclone intensity and size estimation},
  author={Tian, Wei and Zhou, Xinxin and Niu, Xianhua and Lai, Linhong and Zhang, Yonghong and Sian, Kenny Thiam Choy Lim Kam},
  journal={IEEE Journal of Selected Topics in Applied Earth Observations and Remote Sensing},
  volume={16},
  pages={1057--1071},
  year={2022},
  publisher={IEEE}
}

@incollection{Yan2024,
  title={Phy-CoCo: Physical Constraint-Based Correlation Learning for Tropical Cyclone Intensity and Size Estimation},
  author={Yan, Hanting and Mu, Pan and Huang, Cheng and Zhang, Jinglin and Bai, Cong},
  booktitle={ECAI 2024},
  pages={2226--2233},
  year={2024},
  publisher={IOS Press}
}

@inproceedings{zhou2023,
  title={A physics-guided nn-based approach for tropical cyclone intensity estimation},
  author={Zhou, Ziheng and Zhao, Ying and Qing, Yiyu and Jiang, Wenming and Wu, Yihan and Chen, Wenguang},
  booktitle={Proceedings of the 2023 SIAM International Conference on Data Mining (SDM)},
  pages={388--396},
  year={2023},
  organization={SIAM}
}

@article{Huang2022,
  title={MMSTN: A Multi-Modal Spatial-Temporal Network for Tropical Cyclone Short-Term Prediction},
  author={Huang, Cheng and Bai, Cong and Chan, Sixian and Zhang, Jinglin},
  journal={Geophysical Research Letters},
  volume={49},
  number={4},
  pages={e2021GL096898},
  year={2022},
  publisher={Wiley Online Library}
}

@inproceedings{Huang2023,
  title={MGTCF: multi-generator tropical cyclone forecasting with heterogeneous meteorological data},
  author={Huang, Cheng and Bai, Cong and Chan, Sixian and Zhang, Jinglin and Wu, YuQuan},
  booktitle={Proceedings of the AAAI Conference on Artificial Intelligence},
  volume={37},
  number={4},
  pages={5096--5104},
  year={2023}
}

@article{kumar2023,
  title={Tropical cyclone intensity and track prediction in the bay of Bengal using LSTM-CSO method},
  author={Kumar, J Senthil and Venkataraman, V and Meganathan, S and Krithivasan, Kannan},
  journal={IEEE Access},
  volume={11},
  pages={81613--81622},
  year={2023},
  publisher={IEEE}
}

@inproceedings{chen2021,
  title={Real-time tropical cyclone intensity estimation by handling temporally heterogeneous satellite data},
  author={Chen, Boyo and Chen, Buo-Fu and Chen, Yun-Nung},
  booktitle={Proceedings of the AAAI conference on artificial intelligence},
  volume={35},
  number={17},
  pages={14721--14728},
  year={2021}
}

@inproceedings{Nagarajan2014,
  title={Design of a cubesat computer architecture using COTS hardware for terrestrial thermal imaging},
  author={Nagarajan, Chandrasekhar and D'souza, Rodney Gracian and Karumuri, Sukumar and Kinger, Krishna},
  booktitle={2014 IEEE International Conference on Aerospace Electronics and Remote Sensing Technology},
  pages={67--76},
  year={2014},
  organization={IEEE}
}

@article{Altinok2016,
  title={Real-Time Orbital Image Analysis Using Decision Forests, with a Deployment Onboard the IPEX Spacecraft},
  author={Altinok, Alphan and Thompson, David R and Bornstein, Benjamin and Chien, Steve A and Doubleday, Joshua and Bellardo, John},
  journal={Journal of Field Robotics},
  volume={33},
  number={2},
  pages={187--204},
  year={2016},
  publisher={Wiley Online Library}
}

@article{Giuffrida2021,
  title={The $\Phi$-Sat-1 mission: The first on-board deep neural network demonstrator for satellite earth observation},
  author={Giuffrida, Gianluca and Fanucci, Luca and Meoni, Gabriele and Bati{\v{c}}, Matej and Buckley, L{\'e}onie and Dunne, Aubrey and Van Dijk, Chris and Esposito, Marco and Hefele, John and Vercruyssen, Nathan and others},
  journal={IEEE Transactions on Geoscience and Remote Sensing},
  volume={60},
  pages={1--14},
  year={2021},
  publisher={IEEE}
}

@article{Park2020,
  title={Rgb image prioritization using convolutional neural network on a microprocessor for nanosatellites},
  author={Park, Ji Hyun and Inamori, Takaya and Hamaguchi, Ryuhei and Otsuki, Kensuke and Kim, Jung Eun and Yamaoka, Kazutaka},
  journal={Remote Sensing},
  volume={12},
  number={23},
  pages={3941},
  year={2020},
  publisher={MDPI}
}

@article{Pitonak2022,
  title={Cloudsatnet-1: Fpga-based hardware-accelerated quantized cnn for satellite on-board cloud coverage classification},
  author={Pitonak, Radoslav and Mucha, Jan and Dobis, Lukas and Javorka, Martin and Marusin, Marek},
  journal={Remote Sensing},
  volume={14},
  number={13},
  pages={3180},
  year={2022},
  publisher={MDPI}
}

@inproceedings{Oktaviani2023,
  title={The development of experimental remote sensing cubesat payload integrated with on-board classification feature: the progress and educational aspect},
  author={Oktaviani, Shindi Marlina and Santoso, Aipujana T and Abbas, Yasir MO and Purio, Mark Angelo C and Mardiansyah, Galuh and others},
  booktitle={IGARSS 2023-2023 IEEE International Geoscience and Remote Sensing Symposium},
  pages={253--256},
  year={2023},
  organization={IEEE}
}

@inproceedings{Bahl2022,
  title={Scanner neural network for on-board segmentation of satellite images},
  author={Bahl, Ga{\'e}tan and Lafarge, Florent},
  booktitle={IGARSS 2022-2022 IEEE International Geoscience and Remote Sensing Symposium},
  pages={3504--3507},
  year={2022},
  organization={IEEE}
}

@article{Liu2024,
  title={Kan: Kolmogorov-arnold networks},
  author={Liu, Ziming and Wang, Yixuan and Vaidya, Sachin and Ruehle, Fabian and Halverson, James and Solja{\v{c}}i{\'c}, Marin and Hou, Thomas Y and Tegmark, Max},
  journal={arXiv preprint arXiv:2404.19756},
  year={2024}
}

@article{Bessho2016,
  title={An introduction to Himawari-8/9—Japan’s new-generation geostationary meteorological satellites},
  author={Bessho, Kotaro and Date, Kenji and Hayashi, Masahiro and Ikeda, Akio and Imai, Takahito and Inoue, Hidekazu and Kumagai, Yukihiro and Miyakawa, Takuya and Murata, Hidehiko and Ohno, Tomoo and others},
  journal={Journal of the Meteorological Society of Japan. Ser. II},
  volume={94},
  number={2},
  pages={151--183},
  year={2016},
  publisher={Meteorological Society of Japan}
}

@article{Knapp2010,
  title={The international best track archive for climate stewardship (IBTrACS) unifying tropical cyclone data},
  author={Knapp, Kenneth R and Kruk, Michael C and Levinson, David H and Diamond, Howard J and Neumann, Charles J},
  journal={Bulletin of the American Meteorological Society},
  volume={91},
  number={3},
  pages={363--376},
  year={2010},
  publisher={American Meteorological Society}
}

@misc{Gahtan2024,
  author = {Gahtan, J. and Knapp, K. R. and Schreck, C. J. and Diamond, H. J. and Kossin, J. P. and Kruk, M. C.},
  title = {International Best Track Archive for Climate Stewardship ({IBTrACS}) Project, Version 4r01},
  howpublished = {NOAA National Centers for Environmental Information},
  year = {2024},
  doi = {10.25921/82ty-9e16},
  note = {[indicate subset used]}
}

@article{huang2025,
  title={Benchmark dataset and deep learning method for global tropical cyclone forecasting},
  author={Huang, Cheng and Mu, Pan and Zhang, Jinglin and Chan, Sixian and Zhang, Shiqi and Yan, Hanting and Chen, Shengyong and Bai, Cong},
  journal={Nature Communications},
  volume={16},
  number={1},
  pages={5923},
  year={2025},
  publisher={Nature Publishing Group UK London}
}

@article{Pu2025,
  title={A fast physics-based perturbation generator of machine learning weather model for efficient ensemble forecasts of tropical cyclone track},
  author={Pu, Jingchen and Mu, Mu and Feng, Jie and Zhong, Xiaohui and Li, Hao},
  journal={npj Climate and Atmospheric Science},
  volume={8},
  number={1},
  pages={128},
  year={2025},
  publisher={Nature Publishing Group UK London}
}

@article{Nandal2025,
  title={Tropical cyclone intensity estimation based on YOLO-NAS using satellite images in real time},
  author={Nandal, Priyanka and Mann, Prerna and Bohra, Navdeep and Aldehim, Ghadah and Elnour, Asma Abbas Hassan and Allafi, Randa},
  journal={Alexandria Engineering Journal},
  volume={113},
  pages={227--241},
  year={2025},
  publisher={Elsevier}
}

@article{Ma2024,
  title={A multiscale and multilayer feature extraction network with dual attention for tropical cyclone intensity estimation},
  author={Ma, Zhaoyang and Yan, Yunfeng and Lin, Jianmin and Ma, Dongfang},
  journal={IEEE Transactions on Geoscience and Remote Sensing},
  volume={62},
  pages={1--15},
  year={2024},
  publisher={IEEE}
}

@article{zhuo2021,
  title={Physics-augmented deep learning to improve tropical cyclone intensity and size estimation from satellite imagery},
  author={Zhuo, Jing-Yi and Tan, Zhe-Min},
  journal={Monthly Weather Review},
  volume={149},
  number={7},
  pages={2097--2113},
  year={2021}
}

@article{chavas2017,
  title={Physical understanding of the tropical cyclone wind-pressure relationship},
  author={Chavas, Daniel R and Reed, Kevin A and Knaff, John A},
  journal={Nature communications},
  volume={8},
  number={1},
  pages={1360},
  year={2017},
  publisher={Nature Publishing Group UK London}
}

@article{xi2020,
  title={Evaluation of a physics-based tropical cyclone rainfall model for risk assessment},
  author={Xi, Dazhi and Lin, Ning and Smith, James},
  journal={Journal of Hydrometeorology},
  volume={21},
  number={9},
  pages={2197--2218},
  year={2020}
}

@article{Wimmers2019,  
 title={Using deep learning to estimate tropical cyclone intensity from satellite passive microwave imagery},
  author={Wimmers, Anthony and Velden, Christopher and Cossuth, Joshua H},
  journal={Monthly Weather Review},
  volume={147},
  number={6},
  pages={2261--2282},
  year={2019}
}

@article{Zhang2019,  
 title={Tropical cyclone intensity estimation using two-branch convolutional neural network from infrared and water vapor images},
  author={Zhang, Rui and Liu, Qingshan and Hang, Renlong},
  journal={IEEE Transactions on Geoscience and Remote Sensing},
  volume={58},
  number={1},
  pages={586--597},
  year={2019},
  publisher={IEEE}
}

@article{Zhang2021,  
 title={Tropical cyclone intensity classification and estimation using infrared satellite images with deep learning},
  author={Zhang, Chang-Jiang and Wang, Xiao-Jie and Ma, Lei-Ming and Lu, Xiao-Qin},
  journal={IEEE Journal of Selected Topics in Applied Earth Observations and Remote Sensing},
  volume={14},
  pages={2070--2086},
  year={2021},
  publisher={IEEE}
}

@article{shang2023,
  title={Faster and lighter meteorological satellite image classification by a lightweight channel-dilation-concatenation net},
  author={Shang, Shuyao and Zhang, Jinglin and Wang, Xing and Wang, Xinghua and Li, Yuanjun and Li, Yuanjiang},
  journal={IEEE Journal of Selected Topics in Applied Earth Observations and Remote Sensing},
  volume={16},
  pages={2301--2317},
  year={2023},
  publisher={IEEE}
}

@article{li2023fault,
  title={A fault diagnosis method based on an improved deep Q-network for the interturn short circuits of a permanent magnet synchronous motor},
  author={Li, Yuanjiang and Wang, Ruiqi and Mao, Runze and Zhang, Yi and Zhu, Kai and Li, Yuanjun and Zhang, Jinglin},
  journal={IEEE Transactions on Transportation Electrification},
  volume={10},
  number={2},
  pages={3870--3887},
  year={2023},
  publisher={IEEE}
}

@article{gao2020learning,
  title={Learning vertex representations for bipartite networks},
  author={Gao, Ming and He, Xiangnan and Chen, Leihui and Liu, Tingting and Zhang, Jinglin and Zhou, Aoying},
  journal={IEEE transactions on knowledge and data engineering},
  volume={34},
  number={1},
  pages={379--393},
  year={2020},
  publisher={IEEE}
}

@article{zhu2022multi,
  title={Multi-granularity episodic contrastive learning for few-shot learning},
  author={Zhu, Pengfei and Zhu, Zhilin and Wang, Yu and Zhang, Jinglin and Zhao, Shuai},
  journal={Pattern Recognition},
  volume={131},
  pages={108820},
  year={2022},
  publisher={Elsevier}
}

@misc{feng20213d,
  title={3d octave and 2d vanilla mixed convolutional neural network for hyperspectral image classification with limited samples. Remote Sens 13 (21)},
  author={Feng, Y and Zheng, J and Qin, M and others},
  year={2021}
}

@article{zhou2021novel,
  title={A novel ground-based cloud image segmentation method by using deep transfer learning},
  author={Zhou, Zecheng and Zhang, Feng and Xiao, Haixia and Wang, Fuchang and Hong, Xin and Wu, Kun and Zhang, Jinglin},
  journal={IEEE Geoscience and Remote Sensing Letters},
  volume={19},
  pages={1--5},
  year={2021},
  publisher={IEEE}
}

@article{zhang2020ensemble,
  title={Ensemble meteorological cloud classification meets internet of dependable and controllable things},
  author={Zhang, Jinglin and Liu, Pu and Zhang, Feng and Iwabuchi, Hironobu and e Ayres, Antonio Artur de H and De Albuquerque, Victor Hugo C and others},
  journal={IEEE Internet of Things Journal},
  volume={8},
  number={5},
  pages={3323--3330},
  year={2020},
  publisher={IEEE}
}

@article{zhang2025s2dbft,
  title={S2DBFT: Spectral-spatial dual-branch fusion transformer for hyperspectral image classification},
  author={Zhang, Yiheng and Wang, Ziqiang and Huang, Meng and Li, Ming and Zhang, Jian and Wang, Shandong and Zhang, Jinglin and Zhang, Heng},
  journal={IEEE Transactions on Geoscience and Remote Sensing},
  year={2025},
  publisher={IEEE}
}

@inproceedings{zhang2012implementation,
  title={Implementation of motion estimation based on heterogeneous parallel computing system with OpenCL},
  author={Zhang, Jinglin and Nezan, Jean-Francois and Cousin, Jean-Gabriel},
  booktitle={2012 IEEE 14th International Conference on High Performance Computing and Communication \& 2012 IEEE 9th International Conference on Embedded Software and Systems},
  pages={41--45},
  year={2012},
  organization={IEEE}
}

@article{miao2021automated,
  title={Automated CCA-MWF algorithm for unsupervised identification and removal of EOG artifacts from EEG},
  author={Miao, Minmin and Hu, Wenjun and Xu, Baoguo and Zhang, Jinglin and Rodrigues, Joel JPC and De Albuquerque, Victor Hugo C},
  journal={IEEE Journal of Biomedical and Health Informatics},
  volume={26},
  number={8},
  pages={3607--3617},
  year={2021},
  publisher={IEEE}
}

@article{li2021clothing,
  title={Clothing sale forecasting by a composite GRU--Prophet model with an attention mechanism},
  author={Li, Yuanjiang and Yang, Yi and Zhu, Kai and Zhang, Jinglin},
  journal={IEEE Transactions on Industrial Informatics},
  volume={17},
  number={12},
  pages={8335--8344},
  year={2021},
  publisher={IEEE}
}
\appendix
\end{document}